\documentclass{eeict}
\inputencoding{cp1250}
\usepackage{caption3}
\usepackage{float}
\usepackage{hyperref}
\usepackage{multirow}
\usepackage[labelfont=bf]{caption}

\title{SIFT and SURF based feature extraction for the anomaly detection}
\author{Simon Bilik}
\programme{Doctoral Degree Programme (3), FEEC BUT}
\emails{bilik@feec.vutbr.cz}

\supervisor{Karel Horak}
\emailv{horak@feec.vutbr.cz}

\abstract{\sloppy In this paper, we suggest a way, how to use SIFT and SURF algorithms to extract the image features for anomaly detection. We use those feature vectors to train various classifiers on a real-world dataset in the semi -supervised (with a small number of faulty samples) manner with a large number of classifiers and in the one-class (with no faulty samples) manner using the SVDD and SVM classifier. We prove, that the SIFT and SURF algorithms could be used as feature extractors, that they could be used to train a semi-supervised and one-class classifier with an accuracy around 89\% and that the performance of the one-class classifier could be comparable to the semi-supervised one. We also made our dataset and source code publicly available.}

\begin{document}

\maketitle

\selectlanguage{english}
		
		\section{Introduction}
		The anomaly detection (AD) techniques are state-of the art methods for outliers detection in industrial, or medical diagnosis tasks. In comparison with the classification techniques, they are more robust to the unknown samples and they usually require only a small amount of anomalous data to train. This could be advantageous if the anomalous data are difficult to obtain.\\[4.8pt]
		Those algorithms usually consist of a pre-trained convolutional neural network (CNN) to extract image features followed by various one-class classifiers (OCC) to make decision on the input data. CNNs are fast and powerful state of the art methods with a high generalization ability of the input samples. On the other hand, their decision analysis is often complicated and not as clear as the traditional computer vision methods. For those reasons, we describe an experiment, where we use SIFT and SURF algorithms to extract image features in order to process them with selected classifiers.
		\section{Related research}
		Feature points detectors such as SIFT \cite{SIFT}, or the similar and faster SURF \cite{SURF}, are widely used in the area of image matching, or image recognition. These methods search for significant points (usually corners) in the input image and they are invariant to geometric transformations, to light conditions, or image noise. Every feature point is uniquely defined by its high dimensional feature vector, which describes its surroundings. \\[4.8pt]
		As stated above, these methods are also used for object recognition. In its simplest form, it consists of comparing the obtained feature points from the input image with the database of descriptors and finding its nearest neighbour with numerous approaches to the efficient search, indexing and suppressing of false matches. Nevertheless, these often fail on obtaining generic image features and therefore are not so suitable to detect general object categories \cite{SIFT}.\\[4.8pt]
		An example of the application of the feature point based object detection in the traffic domain is shown for example in \cite{Horak2016}, or in \cite{Horak2017}. In the \cite{Horak2016}, the authors use various feature detectors as Harris, FAST, or SURF to solve the Licence Plate Recognition (LPR) problem. After application of the selected detector to the input image, the licence plate is found based on its known dimensions, known orientation of the car and feature points locations. The latter article \cite{Horak2017} uses more similar approach to the one described in \cite{SIFT}, where the feature points obtained from the input image are classified as potential corners using a pre-trained model as LDA, SVM, or a decision tree.
		A number of articles also use the feature points detectors for the AD task - such as \cite{AD_GPS}, where the authors use SURF algorithm as a part of the processing pipeline, or \cite{Sedik2019}, \cite{AD_Crowd} and \cite{AD_Vehicle}, which are based on the SIFT algorithm. The authors of \cite{Sedik2019} introduce a method to recognize epileptic seizures based on the EEG signal processing. The signal is reshaped into 2D space and then is transformed to the frequency domain using FFT. The SURF algorithm is applied to the amplitude spectrum and the classification itself is performed via statistical analysis of the SURF points in time and frequency domain. The results prove a promising potential of this approach, especially in the frequency domain.\\[4.8pt]
		In the paper \cite{AD_Crowd}, the authors use SIFT descriptors to track people in the video surveillance frames and to estimate their motion in order to recognize potentially dangerous behavior. Unfortunately, the proposed methods are not described very well. Authors of the \cite{AD_Vehicle} present a complex algorithm for the vehicle bottom inspection based on the SIFT descriptors. Captured images are compared to the frames within the specification from the database using standard feature matching techniques and possible defects are detected using the correlation of those two images after their matching. From the paper is not clear how extensive was the dataset and if the proposed algorithm was tested in field.
		\section{Materials and methods}
		\subsection{Dataset description}\label{dataset}
\begin{figure}[H]
            \centering
            \includegraphics[width=12cm,keepaspectratio]{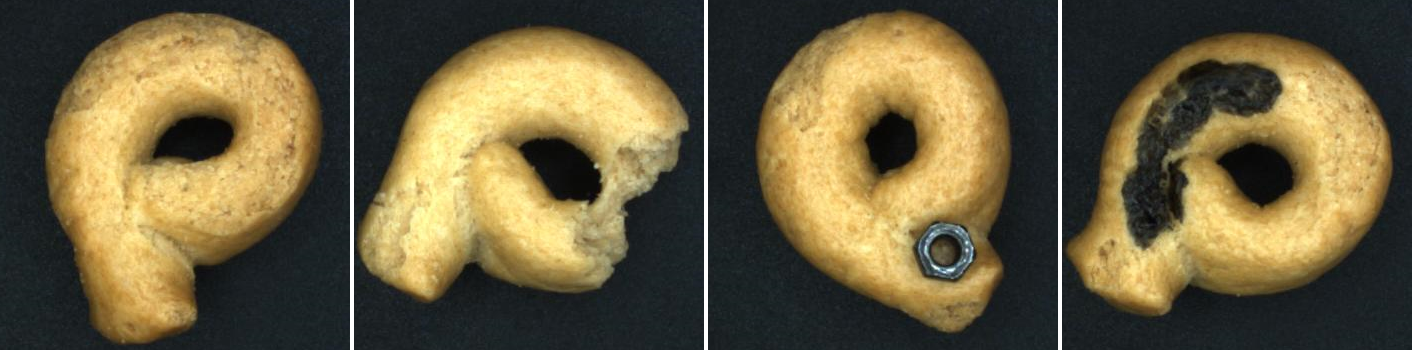}
            \caption{\textit{Cookie Dataset} class samples \cite{Bilik2021}}
            \label{fig:dataset}
        \end{figure}

		For this experiment, we used our internal \textit{Cookie Dataset} described in the previous paper \cite{Bilik2021}. This dataset is designed for anomaly detection task and it captures 1225 original samples in four classes of the \textit{Tarallini} biscuits as can be seen in fig. \ref{fig:dataset}. The whole dataset was augmented by three rotations of 90\textdegree\ to the total of 4900 samples in total and cropped to the size of bounding boxes. Its original structure is as follows:
         \begin{itemize}
            \item No defect (474 captions)
            \item Defect: not complete (465 captions)
            \item Defect: strange object (158 captions)
            \item Defect: color defect (128 captions)
         \end{itemize}
         For the purpose of the anomaly detection experiment, we consider only two classes - the samples within specification (OK class) and randomly selected faulty samples from all faulty classes (NOK class). We created two training datasets - a semi-supervised one with 1000 OK samples and 50 NOK samples as the training set for the first experiment and an one-class one with the same number of OK samples and with no NOK samples. The test dataset contains 200 OK samples and 200 NOK samples and it was used for both experiments. The sample ratio of the NOK classes in the test set was set as 0,4 : 0,3 : 0,3 (not complete : strange object : color defect).
		\subsection{Experiment description}
\begin{figure}[H]
            \centering
          \includegraphics[width=12cm,keepaspectratio]{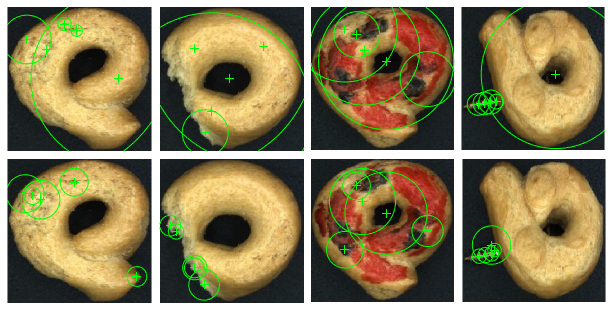}
          \caption{SIFT (upper half) and SURF (lower half) features of the OK (first column) an NOK samples}
          \label{fig:features}
        \end{figure}

	    In this experiment, we use SIFT and SURF algorithms for the feature extraction, as a first part of the AD pipeline, together with semi-supervised and one-class methods for the samples classification. We assume, that the NOK samples have stronger and differently distributed significant points, than the OK ones because of the strange objects presence, or due to various defects. An example illustrating this concept is shown at fig. \ref{fig:features}, where we can see a significant difference in the point's scale and distribution especially between the OK samples and NOK samples with color defect and strange object.
	    At first, we tried to directly use the SIFT and SURF descriptors of five strongest feature points to create a descriptor vectors of each sample, but they had not proved as strong descriptors. For this reason, we decided to use the scale and metric properties of five strongest feature points as the descriptor of \textit{dim(1, 10)}. When extracting the five strongest features, a smaller number of the feature points was detected on a low number of input samples, which is possibly caused by a not distinct surface of the OK samples. In those cases, the missing features were set to zero. Input data were normalized using the z-score for the OCC experiments.\\[4.8pt]
	    As a classifier, we compare the performance of the semi-supervised and one-class approach. In the first case, all models were trained using a semi-supervised dataset with 5\% of NOK samples without further distinguishing between its classes and various classifiers from the Matlab \textit{Classification Learner} toolbox using the five cross-validation, from these we chose three models with the best test accuracy.\\[4.8pt]
	    One-class experiment was performed with the Support Vector Data Description (SVDD) classifier with the implementation available from \cite{MatlabSVDD} and with the standard Matlab Support Vector Machine (SVM) classifier. For further analysis, we trained these models with both semi-supervised and one-class datasets. In the case of OC-SVM, optimal decision boundary of the anomaly score was determined using the test dataset.\\[4.8pt]
	    For the evaluation and comparison of the experiment results, we decided to use the test accuracy and the test AUC (area under the ROC curve) metric, which tells us more precisely about the quality of classification together with the false and true positive rates than the test accuracy. The higher the AUC is, the steeper is the ROC curve and the better the classification results are.\\[4.8pt]
	    The whole experiment was programmed in Matlab 2021b using our own scripts and the Matlab \textit{Classification Learner} toolbox. For the dataset parser, a Python script was used , which is available together with the Matlab code at \cite{BilikGit}.
		\subsection{Experiment results}
		An overview of the model with the highest test accuracy of the semi-supervised experiment is shown in table \ref{table_I}. During the training phase, almost all examined models reached an accuracy over 95\% and a high training AUC. Those metrics significantly fell during the testing, but the results still seem promising and show, that even a small number of a NOK samples is sufficient to learn a classifier. The SURF feature extractor outperforms the SIFT one in both inference time and accuracy results, where it has approximately 10\% higher accuracy and three-times shorter inference time, than the SIFT. Performance of the best trained models is similar in both cases.
		\begin{table}[h]
			\caption{Results of the semi-supervised experiment}
			\vspace*{-3mm}
			\renewcommand*\footnoterule{}
			\centering
			\begin{tabular}{p{0.2\textwidth}p{0.3\textwidth}p{0.2\textwidth}p{0.15\textwidth}}
				\hline\hline
				\textbf{Feature~extractor} &    \textbf{Model} &    \textbf{Test~accuracy~[\%]} & \textbf{AUC~-~Test}\\ 
				\hline
				\multirow{3}{4em}{SIFT} &  Logistic~Regression &    77,2 &   0.86\\
				 &  Coarse~Tree &    77,0 &   0.83\\ 
				 &  One-layered NN &    75,5 &   0.83\\
				 \hline
				\multirow{3}{4em}{SURF} &  Naive~Bayes~(Gaussian) &    89,0 &   0.92\\
				 &  Subspace~kNN &    88,8 &   0.90\\ 
				 &  RUSBoosted Trees &    88,0 &   0.91\\ 
				\hline
			\end{tabular}
			\label{table_I}
		\end{table}
		A comparison of the models trained during the one-class experiment is shown in table \ref{table_II}. Similarly as in the previous case, the metrics showed very good results during the training, but they significantly fell during the testing phase. Both classificators show only slightly better results for the semi-supervised training and the SURF feature extractor clearly outperforms the SIFT while using the OC-SVM classifier. The SVDD classifier didn't prove well in this case and the results of the OC-SVM algorithm with the SURF feature extractor are comparable to the best results achieved with the semi-supervised experiment.
		\begin{table}[h]
			\caption{Results of the one-class experiment}
			\vspace*{-3mm}
			\renewcommand*\footnoterule{}
			\centering
			\begin{tabular}{p{0.2\textwidth}p{0.3\textwidth}p{0.2\textwidth}p{0.15\textwidth}}
				\hline\hline
				\textbf{Feature~extractor} &    \textbf{Model} &    \textbf{Test~accuracy~[\%]} & \textbf{AUC~-~Test}\\ 
				\hline
				\multirow{3}{4em}{SIFT} &  One-class SVDD &    60,0 &   0,60\\
				 &  One-class SVM &    75,7 &   0,80\\ 
				 &  Semi-supervised SVDD &    63,7 &    0,63\\
				 &  Semi-supervised SVM &    79,2 & 0,79\\
				 \hline
				\multirow{3}{4em}{SURF} &  One-class SVDD &    56,0 &   0,56\\
				 &  One-class SVM &    87,5 &   0,92\\ 
				 &  Semi-supervised SVDD &    60,0 &    0,6\\
				 &  Semi-supervised SVM &    89,5 & 0,90\\
				 \hline
			\end{tabular}
			\label{table_II}
		\end{table}
		\section{Discussion}
		The results of both experiments show, that SIFT and SURF algorithms could be used as a feature extractors for anomaly detection in both semi-supervised and one-class manner. In the semi-supervised experiment and using OC-SVM classifier, the SIFT algorithm was clearly outperformed by the SURF in test accuracy, AUC and inference time. Further experiments could be performed to evaluate the minimal number of the NOK samples required in the training dataset to successfully train the model. The best trained models were the Logistic Regression for the SIFT algorithm with test accuracy of 77,2\% and the Naive Bayes for the SURF algorithm with test accuracy of 89,0\%.\\[4.8pt]
		In the one-class experiment, the best results were achieved with the SURF algorithm together with the SVM classifiers reaching the test accuracy of 87,5\% using the One-class training and 89,5\% usign the Semi-supervised approach. For the SVDD classifiers, the SIFT algorithm proved to be a better feature extractor than SURF. The results between the training on the semi-supervised and one-class dataset are not such significant and we can assert, that one-class learning with the SIFT and SURF features is possible. The OCC results achieved with the SURF feature extractor and SVM classifier are comparable to the results of the semi-supervised experiment.
		\section{Conclusion}
		In this paper, we used the SIFT and SURF algorithms as feature extractors for anomaly detection. For the evaluation of the experiment, we compared a semi-supervised learning approach with a small number of a faulty samples and a one-class approach with no faulty samples used during the classifier training. The results clearly show the potential of this approach, when we reached a test accuracy of around 89\% and AUC around 0.90 with the both semi-supervised and one-class learning. The results of the semi-supervised and one-class learning are comparable, so we can say that the semi-supervised learning has not brought a significant improvement in the second part of experiment, although it could be used to train a conventional classifier as was shown in the first part of the experiment.\\[4.8pt]
		In this paper, we prove, that the SIFT and SURF algorithms designed for detection of the image's significant points could be successfully used instead of the convolutional neural networks for the feature extraction and we suggest a way to build a descriptor vector from the obtained parameters. To the best of our knowledge, we are the first to use the SIFT and SURF feature extractors to create a descriptor vectors for the classification, or the anomaly detection task.\\[4.8pt]
		As a future development, we plan to test our method on another anomaly detection dataset and to test it with the preprocessing algorithms based on the convolutional and variational autoencoders presented in our previous work \cite{Bilik2021}. We also plan on further developing the proposed feature extraction method in order to use more parameters  given by the SIFT and SURF algorithms and also to test it with more one-class classificators.\\[4.8pt]
		We made our code and dataset publicly available at \cite{BilikGit} to allow other researchers to quickly reproduce the achieved results and to test our proposed method on their own dataset.
		\section*{Acknowledgement}
		The completion of this paper was made possible by the grant No. FEKT-S-20-6205 -"Research in Automation, Cybernetics and Artificial Intelligence within Industry 4.0" financially supported by the Internal science fund of Brno University of Technology.\\[4.8pt]

\end{document}